\documentclass[11pt]{article}
\usepackage[utf8]{inputenc}
\usepackage[T1]{fontenc}
\usepackage{graphicx}
\usepackage{amsmath,amssymb,amsfonts}
\usepackage{amsthm}
\usepackage{booktabs}
\usepackage{hyperref}

\begin{document}
	
	\title{Grounding-Aware Judge Reward Reinforcement Learning for Out-of-Domain Face Anti-Spoofing}
	
	\author{
		Xiaoyong Yu\textsuperscript{1,2}\\
		Xinge Yu\textsuperscript{1,*}
	}
	
	\date{}
	
	\maketitle
	
	\noindent\textsuperscript{1}School of Electronic Information and Communications, Huazhong University of Science and Technology, Wuhan, Hubei 430074, China\\
	\textsuperscript{2}Mashang Consumer Finance Co., Ltd., Chongqing 400000, China\\
	\textsuperscript{*}Corresponding author: youxg@mail.hust.edu.cn
	
	\begin{abstract}
Face Anti-Spoofing (FAS) must distinguish genuine faces from presentation attacks. A unified Multimodal Large Language Model (MLLM) is an attractive solution: it detects physical attacks (print, replay, mask) in one model and produces interpretable forensic evidence via Chain-of-Thought (CoT). A central obstacle to post-training such a model with reinforcement learning is that FAS reasoning has no rule-verifiable ground truth: only the binary verdict is checkable, while the reasoning chain cannot be verified by any hand-crafted rule. Optimizing the verdict alone is therefore a sparse, hackable signal. The model can reach the correct answer through a shortcut such as a false-rejection bias, which does not generalize. We address this with a grounding-aware judge reward. A 32B vision-language judge scores each reasoning chain against the image on evidence anchoring, logical coherence, and domain knowledge, turning the image itself into the verification signal. The reward is applied in a Group Relative Policy Optimization (GRPO) stage after CoT supervised fine-tuning. It is the judge, rather than the RL objective, that supplies the process supervision a verdict-only reward cannot. Our study starts from a model that already performs well In-Domain (ID) after CoT-based supervised fine-tuning and asks how to improve its Out-of-Domain (OOD) generalization. We find that grounding-aware rewards are an effective means to that end. We validate the method on OOD generalization, where the reward reduces ACER on two physical benchmarks (LCC-FASD 0.288 to 0.273, MSU-MFSD 0.431 to 0.409) while preserving ID accuracy (CelebA-Spoof 0.033 to 0.035). Against ResNet-50/101 and CLIP-ViT-B/16 baselines, the MLLM dominates ID and matches or surpasses the strongest baseline OOD. To our knowledge this is the first use of LLM-as-judge process supervision for FAS.

\end{abstract}
	
	\noindent\textbf{Keywords:} Face anti-spoofing, Multimodal large language model, Reinforcement learning, LLM-as-a-judge, Out-of-domain generalization
	
	\section{Introduction}\label{sec1}
	Face Anti-Spoofing (FAS) is the task of distinguishing genuine faces from presentation attacks, and it serves as a critical first line of defense for face recognition systems deployed in unconstrained real-world settings. The attacks it must counter fall into two broad classes: physical attacks, which re-present a real face to the camera through a print, a replayed video, or a mask, and digital attacks, such as deepfakes, which manipulate pixels directly; this paper focuses on physical attacks. FAS is evaluated on dedicated benchmarks such as CelebA-Spoof~\cite{zhang2020celeba}, LCC-FASD~\cite{liu2018learning}, and MSU-MFSD~\cite{wen2015msu}, and recent surveys provide a comprehensive overview of the field~\cite{yu2023survey}. The central challenge is cross-domain generalization. Presentation artifacts are inherently fine-grained: the cues that separate a bona fide face from an attack, such as moir\'e patterns, specular highlights, printing texture, and subtle depth discontinuities, are weak and easily confounded by benign factors. These cues also vary substantially with the capture device, the illumination conditions, and the specific attack instrument used to fabricate the presentation. Consequently, a model that performs well ID often degrades sharply on unseen attack domains, and closing this generalization gap remains the principal obstacle to deploying FAS reliably in practice.
	
\begin{figure}[htbp]
		\centering
		\includegraphics[width=\textwidth]{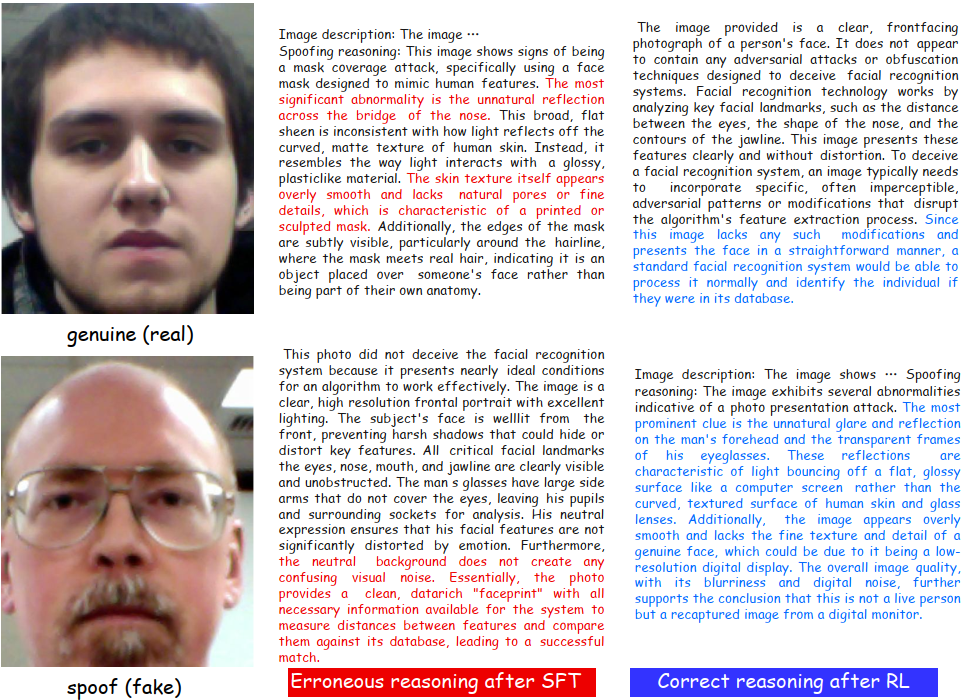}
		\caption{Qualitative examples on genuine faces where CoT SFT falsely rejects the face (hallucinating glare/reflection as a spoof cue) while the grounding-aware RL model correctly grounds its verdict in the actual image content.}
		\label{fig:case}
	\end{figure}

	MLLMs~\cite{qwen2025vl,radford2021clip} offer a unified FAS paradigm: a single model detects attacks across types~\cite{jiang2025ms,he2024joint,ma2026pafas} and produces interpretable forensic evidence in natural language. This paradigm has already shown concrete promise. Face-CoT introduces the first large-scale Visual Question Answering (VQA) dataset tailored for FAS, covering 14 spoofing attack types with high-quality CoT annotations and demonstrates that MLLMs trained with CoT-enhanced progressive learning outperform prior state-of-the-art methods on multiple benchmarks. CoT supervision teaches the model to verbalize the physical artifacts, e.g., a screen frame, a moir\'e pattern, a specular reflection, and a paper texture, yielding strong ID accuracy and an auditable reasoning path. However, supervised fine-tuning alone does not explicitly optimize the reasoning process, and it is not obvious how reinforcement learning should be applied when the reasoning has no ground truth. More fundamentally, Supervised Fine-Tuning (SFT) tends to induce visual hallucinations under distribution shifts, which is precisely the failure mode that cross-domain FAS cannot afford: the model may latch onto spurious textual patterns that correlate with labels in the training domain rather than learning to ground its verdict in the actual image evidence. This risk is compounded by the intrinsic difficulty of the task itself: as illustrated in Fig.~\ref{fig:case}, some attack samples are so subtle that even the human eye struggles to tell them apart from genuine faces, leaving the model with only faint, easily confounded cues to rely on. Reinforcement learning offers a bidirectional reward-driven mechanism with iterative error correction that can refine visual perception and discourage such shortcut reliance. Nevertheless, applying RL to FAS reasoning is impeded by a fundamental obstacle: because the reasoning chain admits no rule-verifiable ground truth, a verdict-only reward is inherently sparse and susceptible to reward hacking. It is this gap that motivates a grounding-aware reward signal capable of supervising the reasoning process itself.
	
	Reinforcement Learning has transformed reasoning in math and code~\cite{ouyang2022instructgpt,lightman2023let,shao2024deepseekmath}, where the answer is exactly verifiable and a verifiable reward lets the model re-rank its reasoning paths toward correct ones. FAS does not share this property: only the final binary verdict is checkable, and the reasoning chain has no ground truth. This is the verification gap that blocks RL post-training for FAS. Optimizing the verdict alone is a sparse, hackable signal. The model can reach the correct answer through a shortcut such as a false-rejection spoof bias, which does not generalize to unseen domains. Our key observation is that, although FAS reasoning has no rule-verifiable ground truth, it does have an image-grounded one: a correct reasoning chain must describe evidence that is actually present in the image. We turn this observation into a reward. A vision-language judge scores each reasoning chain against the image on evidence anchoring, logical coherence, and domain knowledge, and this grounding-aware judge reward is applied in a GRPO stage after SFT. Because physical artifacts are verbalizable and image-grounded, the judge can verify whether the reasoning is anchored in the actual image. A signal that hand-crafted rules cannot provide. The method is thus a direct answer to the verification gap: where RLVR needs a rule, we substitute a judge grounded in the image.
	
	We validate the method on OOD generalization, which is the setting where the difference between grounded and shortcut reasoning becomes observable. The grounding-aware judge reward reduces OOD ACER on two physical benchmarks while preserving ID accuracy. A full ablation further isolates the judge reward as the ingredient that extends the gain to the harder benchmark rather than the RL objective. We further compare the MLLM against Convolutional Neural Network (CNN) and Contrastive Language-Image Pre-training (CLIP) baselines and position our method relative to recent rule-based RL-FAS approaches~\cite{ma2026pafas,wang2026fasr1,zhang2026tarfas}. Our contributions are twofold. (ii) We validate the method on OOD generalization against CNN and CLIP baselines and position it relative to recent rule-based RL-FAS approaches. To our knowledge this is the first use of LLM-as-judge process supervision for FAS. The remainder of this paper is organized as follows. Section~\ref{sec:related} reviews related work. Section~\ref{sec:method} presents the method. Section~\ref{sec:exp} reports the experiments and ablations. Section~\ref{sec:discussion} discusses the mechanism. Section~\ref{sec:conclusion} concludes.
	
	\section{Related Work}
	\label{sec:related}
	\subsection{Face Anti-Spoofing and Cross-Domain Generalization}
	Classical FAS approaches learn a binary CNN classifier on presentation-attack benchmarks, typically with auxiliary supervision such as depth maps or remote Photoplethysmography (rPPG) signals~\cite{liu2018learning,wen2015msu}. Data-centric analyses of deep FAS further show that the training distribution strongly influences generalization behavior~\cite{sun2021understanding}, and multi-modal feature fusion has been explored as a way to enrich the representation beyond Red-Green-Blue (RGB) appearance~\cite{he2023lightweight}. Because presentation artifacts are fine-grained and domain-specific, cross-domain generalization is the central challenge. A substantial body of work addresses it through domain generalization, test-time adaptation, and domain randomization~\cite{tobin2017domain}. Prompt-based CLIP fine-tuning has recently emerged as a promising direction: by aligning visual features with textual descriptions of spoofing artifacts. These methods achieve stronger generalization than purely visual classifiers~\cite{liu2024cfpl,li2025optimal}. However, they remain discriminative models that output a binary label without an explicit reasoning trace, which limits their interpretability and makes their decisions difficult to audit in security-sensitive deployments.
	
	\subsection{Multimodal Large Language Models for FAS}
	MLLMs have recently been applied to FAS to provide both classification and natural-language explanations. Early work reformulates FAS as a VQA task, where the model is prompted to describe spoofing cues before emitting a verdict~\cite{jiang2025ms}. Subsequent methods extend this paradigm with richer reasoning: FaceShield~\cite{wang2026faceshield} introduces a large-scale instruction dataset for explainable FAS. PA-FAS~\cite{ma2026pafas} augments reasoning paths and shuffles answers to mitigate shortcut learning. FAS-R1~\cite{wang2026fasr1} applies difficulty-aware GRPO to focus optimization on hard attacks. TAR-FAS~\cite{zhang2026tarfas} equips the MLLM with external visual tools for fine-grained forensic investigation. These works establish MLLMs as a viable unified FAS paradigm, but their post-training rewards remain either supervised (SFT-only) or rule-based (consistency checks, format rewards). None verifies whether the reasoning is actually grounded in the image, which is the gap our method fills.
	
	\subsection{Reinforcement Learning Post-Training and Process Rewards}
	Reinforcement Learning from Human Feedback (RLHF) and Reinforcement Learning with Verifiable Rewards (RLVR) have become standard tools for aligning language models~\cite{ouyang2022instructgpt}. GRPO~\cite{shao2024deepseekmath} simplifies Proximal Policy Optimization (PPO) by estimating advantages from group-normalized rewards, eliminating the need for a value network, and has been widely adopted for reasoning tasks. Group Sequence Policy Optimization (GSPO)~\cite{zheng2025gspo} further stabilizes training by operating at the sequence level rather than the token level. A parallel line of work introduces process rewards: rather than scoring only the final answer, a process reward model scores intermediate reasoning steps~\cite{lightman2023let}. LLM-as-judge~\cite{zheng2023judge,yu2025judge} has emerged as a scalable instantiation of this idea, where a strong language model evaluates the quality of a generation. Our grounding-aware judge is a domain-specific instance of this paradigm: it does not score general quality, but specifically whether the FAS reasoning is anchored in the image evidence, a dimension that neither rule-based rewards nor generic judges can evaluate.
	
	\section{Method}
	\label{sec:method}
	
	\subsection{Problem Formulation}
	\label{sec:problem}
	Let $\mathcal{D}=\{(I_i,y_i)\}_{i=1}^{N}$ be a face anti-spoofing dataset, where $I_i$ is a face image and $y_i\in\{\texttt{A},\texttt{B}\}$ is the binary label. A multimodal policy $\pi_\theta$ generates a reasoning chain $c=(c_1,\dots,c_T)$ followed by a verdict $\hat{y}=f(c)\in\{\texttt{A},\texttt{B}\}$. Physical attacks re-present a real face to the camera. Their artifacts are macroscopic and verbalizable: a screen frame, a moir\'e pattern, a specular reflection, a paper texture, a mask boundary.
	
	The standard supervised objective maximizes the likelihood of a reference chain $c^\star$:
	\begin{equation}
		\mathcal{L}_{\mathrm{SFT}}(\theta)=-\mathbb{E}_{(I,c^\star)\sim\mathcal{D}}\Big[\sum_{t=1}^{T}\log\pi_\theta(c_t\mid c_{<t},I)\Big].
		\label{eq:sft}
	\end{equation}
	SFT teaches the model to verbalize forensic evidence but does not explicitly optimize the decision boundary under domain shift~\cite{ma2026pafas,jiang2025exploring}. We therefore seek a post-training objective that improves out-of-domain generalization while preserving in-domain accuracy.
	
	Formally, let $P_{\mathrm{ID}}$ and $P_{\mathrm{OOD}}$ denote the in-domain and out-of-domain image distributions. The goal is to find $\theta$ that minimizes the expected error on $P_{\mathrm{OOD}}$,
	\begin{equation}
		\min_{\theta}\ \mathbb{E}_{(I,y)\sim P_{\mathrm{OOD}}}\big[\mathbb{1}(f(c)\neq y)\big]\quad\text{s.t.}\quad \mathbb{E}_{(I,y)\sim P_{\mathrm{ID}}}\big[\mathbb{1}(f(c)\neq y)\big]\leq\epsilon.
		\label{eq:ood}
	\end{equation}
	
	\subsection{Why Verdict-Only Rewards Are Insufficient}
	\label{sec:why}
	A natural RL objective rewards only the final verdict:
	\begin{equation}
		r_{\mathrm{verdict}}(c,y)=\mathbb{1}(f(c)=y).
		\label{eq:verdict}
	\end{equation}
	This reward is sparse and hackable. Let $\mathcal{C}_y=\{c:f(c)=y\}$. The policy can maximize \eqref{eq:verdict} by concentrating mass on any $c\in\mathcal{C}_y$, including chains whose reasoning is ungrounded. A particularly harmful member is the false-rejection chain, which assigns $\texttt{B}$ to genuine faces via a spurious cue (e.g., glare misread as a spoof artifact). Because such cues correlate with the label on the training distribution but not on $P_{\mathrm{OOD}}$, the policy that exploits them has low OOD accuracy despite high in-domain reward.
	
	Let $\mathcal{G}\subset\mathcal{C}_y$ be the subset of grounded chains. Decompose the expected reward as
	\begin{equation}
		\mathbb{E}_{c\sim\pi}[R(c)]
		=\underbrace{\mathbb{P}_{c\sim\pi}(c\in\mathcal{G})}_{\text{grounding}}
		+\underbrace{\mathbb{P}_{c\sim\pi}(c\in\mathcal{C}_y\setminus\mathcal{G})}_{\text{shortcut}}.
		\label{eq:decomp}
	\end{equation}
	A process reward $r_{\mathrm{proc}}(c,I)$ that scores grounding provides a signal correlated with the first term and orthogonal to the second, reducing the variance of the policy gradient whenever $r_{\mathrm{proc}}$ is positively correlated with $\mathcal{G}$.
	
	\subsection{CoT Supervised Fine-Tuning}
	\label{sec:sft}
	We build on the semantic-level annotation paradigm introduced in our prior work MS-MFAD~\cite{yu2026msmfad}, which constructs a unified benchmark covering 16 categories of physical and digital attacks. For each attack category, 1,000 precise semantic masks are annotated through a dual-person verification and ambiguity arbitration process, yielding reasoning evidence chains that are strictly tied to spoofed regions~\cite{yu2026msmfad}. The resulting physical subset comprises roughly 14K samples, each pairing an image with a structured reasoning chain that follows a fixed template: an image description, followed by \emph{spoofing reasoning} that lists the observed forensic evidence, and a final binary verdict. We fine-tune Qwen2.5-VL-7B~\cite{qwen2025vl} with Low-Rank Adaptation~\cite{hu2021lora} on this physical CoT data. The model emits the chain inside a \texttt{<think>} block and the verdict inside an \texttt{<answer>} block. We freeze the vision encoder and the aligner, and adapt only the LLM backbone with LoRA (rank 8). This yields strong in-domain accuracy (CelebA-Spoof ACER $0.033$), consistent with the 40--50\% relative ACER reduction reported in MS-MFAD under supervised fine-tuning~\cite{yu2026msmfad}.
	
	\subsection{Grounding-Aware Judge Reward RL}
	\label{sec:rl}
\begin{figure}[htbp]
		\centering
		\includegraphics[width=\textwidth]{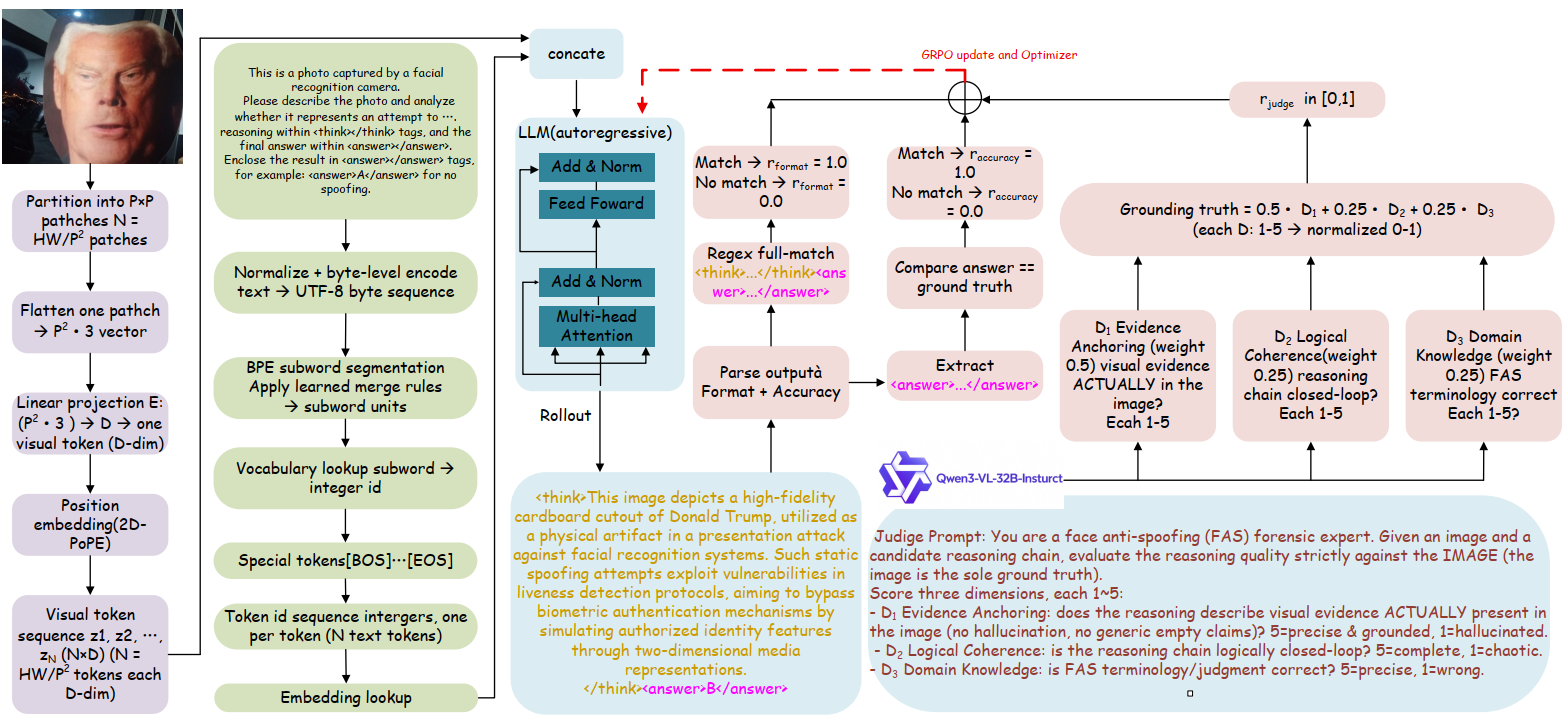}
		\caption{Schematic of the overall reward design. We fine-tune an MLLM with CoT SFT, then apply GRPO with a grounding-aware judge reward that scores reasoning against the image, improving out-of-domain generalization.}
		\label{fig:gr}
	\end{figure}

	Because FAS has no rule-verifiable ground-truth reasoning, we ground the process reward in the image evidence using a 32B vision-language judge $J$. We apply GRPO~\cite{shao2024deepseekmath} with sequence-level importance sampling, following the sequence-level formulation of GSPO~\cite{zheng2025gspo}, and a composite reward. The overall reward design is illustrated in Figure~\ref{fig:gr}.
	
	\subsubsection{Composite Reward}
	\label{sec:reward}
	The composite reward is
	\begin{equation}
		r(c)=w_{\mathrm{fmt}}\,r_{\mathrm{fmt}}(c)+w_{\mathrm{acc}}\,r_{\mathrm{acc}}(c,y)+w_{\mathrm{judge}}\,r_{\mathrm{judge}}(c,I),
		\label{eq:composite}
	\end{equation}
	with $w_{\mathrm{fmt}}=0.1$, $w_{\mathrm{acc}}=0.5$, $w_{\mathrm{judge}}=0.4$. The weights satisfy $w_{\mathrm{acc}}>w_{\mathrm{judge}}>w_{\mathrm{fmt}}$ so that correctness dominates while the judge supplies the process signal.
	
	\subsubsection{Judge Protocol}
	\label{sec:judge}
	The 32B judge scores each chain on three dimensions, each on a 1--5 scale, under the instruction that the image is the sole ground truth. The three dimensions are not chosen arbitrarily: they correspond to three failure modes that FAS reasoning can exhibit, namely evidence hallucination (describing cues absent from the image), logical discontinuity (a missing link between observation and verdict), and domain misjudgment (misusing FAS terminology or decision rules). Together they span the perceptual, inferential, and knowledge levels of the reasoning chain.
	
	\textbf{D1 Evidence Anchoring} asks whether the reasoning describes visual evidence that is actually present in $I$. This dimension targets the most dangerous failure mode of MLLMs under distribution shift: the model may generate plausible-sounding reasoning that is disconnected from the image content. Prior work identifies two systematic failures of MLLM CoT reasoning, a perceptual blind spot and a perception--reasoning disconnect, where the model produces a coherent chain whose content lacks genuine causal grounding in the visual input. In FAS, this disconnect manifests as the model projecting artifact patterns common in the training domain (e.g., screen frame or moir\'e pattern) onto images that do not contain them, producing false accepts or false rejects. D1 turns grounding from an abstract requirement into an operational, scorable dimension by requiring the judge to check each piece of cited visual evidence against the actual image content.
	
	\textbf{D2 Logical Coherence} evaluates whether the chain forms a closed loop from observation through intermediate inference to the final verdict. What is distinctive about FAS reasoning is that even when the evidence is correctly described, the step from observing an artifact to declaring an attack can still be skipped or wrong. For instance, the model may correctly identify a specular highlight yet equate it directly with a screen recapture, ignoring that specular highlights on skin are equally common under natural illumination. PA-FAS~\cite{ma2026pafas} observes that shortcut learning in multimodal FAS reasoning takes the form of mapping the input directly to the answer while bypassing the reasoning process. D2 is designed to detect exactly this bypass: if the chain contains an unstated leap between evidence and conclusion, the coherence score drops, and RL training penalizes such shortcut behavior.
	
	\textbf{D3 Domain Knowledge} examines the accuracy of FAS terminology and the correctness of the judgment rules. This dimension complements D1 and D2: D1 asks what is said (does the evidence exist), D2 asks how it is said (is the reasoning coherent), and D3 asks whether it is said correctly (does the professional judgment hold). For example, the model may correctly describe a texture (high D1) and coherently conclude print attack (high D2), yet mistake paper texture for screen moir\'e, which D3 captures as a knowledge-level error. FAS domain knowledge is highly fine-grained: different attack types (print, replay, mask) differ systematically in their visual cues, and general-purpose vision-language models often lack this fine-grained discriminative ability. D3 lets the judge leverage the visual--linguistic knowledge acquired in pretraining to verify FAS-specific judgments at a professional level.
	
	The judge reward aggregates the three dimensions as
	\begin{equation}
		r_{\mathrm{judge}}(c,I)=\alpha_1\frac{D1-1}{4}+\alpha_2\frac{D2-1}{4}+\alpha_3\frac{D3-1}{4},
		\label{eq:judge}
	\end{equation}
	with $\alpha_1=0.5$ and $\alpha_2=\alpha_3=0.25$. The weighting reflects the relative importance of the three dimensions for supervising FAS reasoning: evidence anchoring receives the largest weight because it is the precondition for the other two to be meaningful; if the cited evidence is absent from the image, then no matter how coherent the logic or how accurate the terminology, the chain has no supervisory value. Logical coherence and domain knowledge each carry a quarter of the weight, jointly constraining the quality of grounded reasoning. Mapping the raw 1--5 scores to $[0,1]$ via $(D-1)/4$ places all dimensions on a common scale before weighting, avoiding implicit bias from differences in scale.
	
	The judge is anchored to the ground-truth label: it evaluates whether the reasoning is consistent with the answer rather than independently re-classifying the image. This design choice is critical. If the judge were asked to also decide whether the image is genuine, then whenever its own classification disagreed with the ground truth, the reward signal would inject noise or even systematic bias. By restricting the judge to ``given the answer, does the reasoning support it?'', we isolate the judge's potential weakness in FAS classification and focus it on what it genuinely excels at: assessing text--image evidence consistency, structural completeness of the logic, and normative correctness of domain statements. In other words, the judge does not answer ``is this image real or fake?'', but ``given the label of this image, does this reasoning honestly, coherently, and professionally arrive at that conclusion?''. This constraint turns LLM-as-judge from a second classifier into a process supervisor, consistent with the core principle of this paper: using the image as the verification signal.
	
	\subsubsection{GRPO Objective with Sequence-Level Importance Sampling}
	\label{sec:grpo}
	For each prompt we sample $N=8$ rollouts. We normalize each rollout's reward by its group mean and standard deviation:
	\begin{equation}
		\hat{A}_i=\frac{r(c_i)-\mathrm{mean}_j\,r(c_j)}{\mathrm{std}_j\,r(c_j)}.
		\label{eq:adv}
	\end{equation}
	Standard GRPO computes a token-level importance ratio and applies clipping per token. However, token-level ratios are computed from a single sample of each next-token distribution and therefore perform no meaningful distribution correction; they inject high-variance noise that accumulates with response length and is amplified by clipping~\cite{zheng2025gspo}. Since the reward is assigned at the sequence level, the correction and clipping should also operate at the sequence level~\cite{zheng2025gspo}. We therefore adopt sequence-level importance sampling following Group Sequence Policy Optimization (GSPO)~\cite{zheng2025gspo}. The length-normalized sequence ratio is:
	\begin{equation}
		s_i(\theta)=\left(\frac{\pi_\theta(c_i\mid I)}{\pi_{\mathrm{ref}}(c_i\mid I)}\right)^{1/|c_i|}
		=\exp\!\left(\frac{1}{|c_i|}\sum_{t=1}^{|c_i|}\log\frac{\pi_\theta(c_{i,t}\mid I,c_{i,<t})}{\pi_{\mathrm{ref}}(c_{i,t}\mid I,c_{i,<t})}\right),
		\label{eq:seqratio}
	\end{equation}
	where $|c_i|$ is the number of tokens in rollout $c_i$. The length normalization is essential: without it, a few tokens' likelihood changes would swing the sequence ratio wildly and different response lengths would require different clipping ranges~\cite{zheng2025gspo}. The policy is updated by maximizing a clipped surrogate objective with a KL anchor toward the SFT reference $\pi_{\mathrm{ref}}$:
	\begin{equation}
		\mathcal{L}(\theta)=\mathbb{E}\Big[\min\big(s_i(\theta)\hat{A}_i,\ \operatorname{clip}(s_i(\theta),1-\epsilon,1+\epsilon)\hat{A}_i\big)-\beta\,D_{\mathrm{KL}}\big(\pi_\theta\,\|\,\pi_{\mathrm{ref}}\big)\Big],
		\label{eq:grpo}
	\end{equation}
	where $\beta=0.1$. Because $s_i(\theta)$ is a length-normalized geometric mean, its numerical scale differs from that of the token-level ratio; we therefore use a correspondingly tighter clipping range following GSPO~\cite{zheng2025gspo}. This formulation weights every token of a rollout equally by the sequence ratio $s_i(\theta)$, rather than by its own noisy token-level ratio, providing a more stable and reliable learning signal~\cite{zheng2025gspo}.
	
	\subsection{Training Details}
	We fine-tune Qwen2.5-VL-7B-Instruct with Low-Rank Adaptation~\cite{hu2021lora} on physical CoT annotations. Each training sample is a triplet of an image, a prompt, and a structured reasoning chain. The reasoning chain follows a fixed two-part template: a natural-language image description of the scene, followed by spoofing reasoning that lists the observed forensic evidence and ends with a binary verdict. The model is trained to emit the whole chain inside a \texttt{<think>} block and the verdict inside an \texttt{<answer>} block, where $\texttt{A}$ denotes a genuine face and $\texttt{B}$ denotes a spoof. The training set consists of $14{,}286$ physical samples from CelebA-Spoof ($9{,}489$ spoof, $4{,}797$ genuine), with $150$ held-out validation samples. The CoT annotations are LLM-assisted structured chains: the description and the forensic reasoning are drafted by a strong MLLM and quality-filtered against the image, so each chain is grounded in the physical attack cues of CelebA-Spoof (print, replay, and mask).
	
	We use a single prompt, shared by the SFT and RL stages: ``This is a photo captured by a facial recognition camera. Please describe the photo and analyze whether it represents an attempt to spoof the facial recognition system. If deception is detected, output B; otherwise, output A. Place your reasoning within \texttt{<think></think>} tags and the final answer within \texttt{<answer></answer>} tags.'' This prompt fixes both the output format (the two tags) and the semantics of the verdict labels, and it is kept unchanged across both stages so that the RL policy and the SFT reference operate on the same input distribution. We freeze the vision encoder and the aligner and adapt only the LLM backbone with LoRA (rank $8$, alpha $32$, dropout $0.05$, applied to all linear layers). We train for $3$ epochs with learning rate $1\times10^{-4}$ (cosine schedule with $5\%$ warm-up), an effective batch size of $32$ (per-device batch $8$ with gradient accumulation $4$), and a maximum sequence length of $2048$ tokens. This yields strong in-domain accuracy (CelebA-Spoof ACER $0.033$).
	
	\section{Experiments}
	\label{sec:exp}
	
	\subsection{Setup}
	We evaluate on one in-domain benchmark (CelebA-Spoof~\cite{zhang2020celeba}) and two out-of-domain benchmarks. Large Crowd-Collected Face Anti-Spoofing Dataset (LCC-FASD)~\cite{liu2018learning} is a large crowd-collected FAS dataset released by Timoshenko et al., organized into training, development, and evaluation splits, and comprising roughly 18.8K files in the public release we use. Michigan State University Mobile Face Spoofing Database (MSU-MFSD)~\cite{wen2015msu} is the MSU Mobile Face Spoofing Database, collected from 35 subjects with two capture devices (an Android phone and a laptop) under two illumination scenes; each subject contributes genuine videos and attacks of three types including printed photo, iPad replay, and iPhone replay. We use the frame-cropped version of MSU-MFSD, in which faces are cropped following the original MATLAB preprocessing script, yielding 77,595 face crops over 279 videos, with no subject overlap between the train and test splits. We report ACER on all benchmarks. OOD generalization is used as the validation setting: it is where the difference between grounded and shortcut reasoning becomes observable. The baselines are ResNet-50, ResNet-101, and CLIP-ViT-B/16~\cite{radford2021clip}, all fine-tuned on the same physical data. Lightweight multi-modal fusion methods~\cite{he2023lightweight} represent an alternative efficiency-oriented direction, but they do not provide the reasoning trace that our method targets. All models are trained and evaluated with ms-swift 4.0.3 under Python 3.12 and CUDA 12.8 in a Singularity container. The 7B policy is trained on one NVIDIA H100 GPU and the 32B judge on a second H100 GPU; evaluation uses a single H100 GPU.
	
	\subsection{Main Comparison}
	\label{sec:main}
	Table~\ref{tab:main} reports the full comparison. We group the results into three blocks: conventional CNN baselines, a CLIP baseline, and the MLLM family (CoT SFT and the grounding-aware judge reward RL). Three observations follow from Table~\ref{tab:main}. First, in-domain, the MLLM family is far ahead of every conventional baseline: CoT SFT reaches ACER $0.033$, roughly a threefold reduction over CLIP-ViT-B/16 ($0.098$) and a fourfold reduction over the ResNet models ($0.132$--$0.134$). This confirms that verbalizing forensic evidence before emitting a verdict is an effective inductive bias for in-domain FAS, and that the gain is not merely a byproduct of model scale, since the same physical training data is used throughout.
	
	\begin{table}[htbp]
		\centering
		\caption{Main comparison. All models are trained on the same physical data and evaluated on one in-domain (ID) and two out-of-domain (OOD) benchmarks. Lower ACER is better; best per column in bold.}
		\label{tab:main}
		\small
		\begin{tabular}{l c c c}
			\toprule
			\textbf{Model} & \textbf{CelebA-Spoof} & \textbf{LCC-FASD} & \textbf{MSU-MFSD} \\
			& (ID) & (OOD) & (OOD) \\
			\midrule
			ResNet-50 & 0.134 & 0.322 & 0.486 \\
			ResNet-101 & 0.132 & 0.334 & 0.487 \\
			CLIP-ViT-B/16~\cite{radford2021clip} & 0.098 & 0.282 & \textbf{0.410} \\
			\midrule
			MLLM CoT SFT (baseline) & \textbf{0.033} & 0.288 & 0.431 \\
			MLLM $+$ judge reward RL (ours) & 0.035 & \textbf{0.273} & 0.409 \\
			\bottomrule
		\end{tabular}
	\end{table}
	
	Second, out-of-domain, the ordering changes. The ResNet baselines degrade markedly ($0.322$--$0.334$ on LCC-FASD and $0.486$--$0.487$ on MSU-MFSD), consistent with the well-documented fragility of purely visual classifiers under capture-device and illumination shift. CLIP-ViT-B/16 generalizes substantially better ($0.282$ on LCC-FASD, $0.410$ on MSU-MFSD), reflecting the transferability of its vision-language representation. CoT SFT sits between the two: it outperforms CLIP in-domain by a large margin but trails it slightly out-of-domain ($0.288$ vs.\ $0.282$ on LCC-FASD, $0.431$ vs.\ $0.410$ on MSU-MFSD). This crossover is the key empirical fact motivating our method: SFT teaches the model to produce fluent, evidence-listing reasoning, but does not explicitly constrain whether that evidence remains anchored in the image once the domain shifts.
	
	Third, applying the grounding-aware judge reward closes exactly this gap. The RL stage reduces OOD ACER on both physical benchmarks (LCC-FASD $0.288\to0.273$, MSU-MFSD $0.431\to0.409$) and, on LCC-FASD, surpasses the strongest conventional baseline ($0.273$ vs.\ $0.282$); on MSU-MFSD it matches CLIP within one point ($0.409$ vs.\ $0.410$) while retaining a nearly threefold in-domain advantage ($0.035$ vs.\ $0.098$). The in-domain cost is negligible ($0.033\to0.035$), i.e., the improvement in generalization is not obtained by trading away in-domain accuracy. The MLLM with the judge reward is thus the only model in Table~\ref{tab:main} that is simultaneously best or near-best in-domain and out-of-domain, which is the property that matters for deployment, where the test domain is unknown in advance. The magnitude of the OOD gain is consistent but modest. We defer the analysis of its size to Section~\ref{sec:discussion}, where we argue that it is bounded by the grounding capability of the judge rather than by the direction of the method.
	
	\subsection{Ablation: The Judge Reward Is the Active Ingredient}
	\label{sec:ablation}
	A natural objection is that any RL stage, not specifically the grounding-aware judge, could produce the OOD gain, for instance through format rewards, accuracy rewards, or hyper-parameter tuning. This section isolates the contribution of each reward component. Table~\ref{tab:judge} ablates the reward composition at a fixed budget of 1000 steps and $\beta{=}0.1$.
	
	\begin{table}[htbp]
		\centering
		\caption{Ablation of the reward composition (all at 1000 steps, $\beta{=}0.1$). Lower ACER is better.}
		\label{tab:judge}
		\small
		\begin{tabular}{l c c c}
			\toprule
			\textbf{Variant} & \textbf{CelebA-Spoof} & \textbf{LCC-FASD} & \textbf{MSU-MFSD} \\
			& (ID) & (OOD) & (OOD) \\
			\midrule
			CoT SFT (baseline) & 0.033 & 0.288 & 0.431 \\
			no-judge (fmt$+$acc) & 0.033 & 0.270 & 0.429 \\
			judge $w{=}0.3$ & 0.029 & 0.281 & 0.417 \\
			judge $w{=}0.5$ & 0.036 & 0.280 & 0.435 \\
			judge $w{=}0.4$, lr $1\times10^{-6}$ & 0.035 & \textbf{0.273} & \textbf{0.409} \\
			\bottomrule
		\end{tabular}
	\end{table}
	
	\textbf{Format and accuracy rewards alone are insufficient.} Removing the judge reward and keeping only the format and accuracy terms (row ``no-judge'') already improves LCC-FASD from $0.288$ to $0.270$, which is even slightly better than the full judge configuration on this benchmark ($0.273$). On MSU-MFSD, however, the same variant is essentially flat ($0.431\to0.429$), whereas the full configuration reaches $0.409$. This asymmetry is informative. LCC-FASD is the closer of the two OOD domains, so a verdict-only signal that still rewards correct classification and well-formed output is enough to yield a modest gain. MSU-MFSD is the harder domain, and there the verdict-only signal saturates: without a process-level constraint on whether the reasoning is anchored in the image, the model has no gradient pushing it away from memorized shortcuts. The judge reward is precisely what extends the gain to the harder benchmark, which is the setting we care about.
	
	\textbf{The judge weight has a broad optimum.} Varying the judge weight between $0.3$ and $0.5$ gives comparable OOD results ($0.281/0.417$ at $w{=}0.3$, $0.280/0.435$ at $w{=}0.5$), while $w{=}0.4$ with a lower learning rate gives the best balance ($0.273/0.409$). Two points are worth noting. At $w{=}0.3$ the in-domain ACER actually improves to $0.029$, suggesting that a moderate amount of process supervision can act as a regularizer that sharpens in-domain decisions; at $w{=}0.5$ the in-domain ACER rises to $0.036$ and MSU-MFSD degrades to $0.435$, indicating that over-weighting the judge lets its (imperfect) grounding assessments dominate the verdict signal and pull the policy away from correct classification. The existence of a broad but non-monotonic optimum is consistent with the judge being a useful but noisy process signal rather than an oracle.
	
	\textbf{The active ingredient is the judge, not the RL machinery.} The no-judge row shares every other component with the full configuration (same policy, same GRPO objective, same rollout budget, same KL anchor) and differs only in the reward term. Its failure to improve the harder benchmark therefore isolates the judge reward as the cause of the OOD gain, rather than the RL objective, the sampling procedure, or the optimization schedule. Section~\ref{sec:discussion} reinforces this conclusion with a hyper-parameter sweep, where no setting of the learning rate or KL coefficient recovers the MSU-MFSD gain without the judge reward.
	
	\begin{figure}[htbp]
		\centering
		\includegraphics[width=\textwidth]{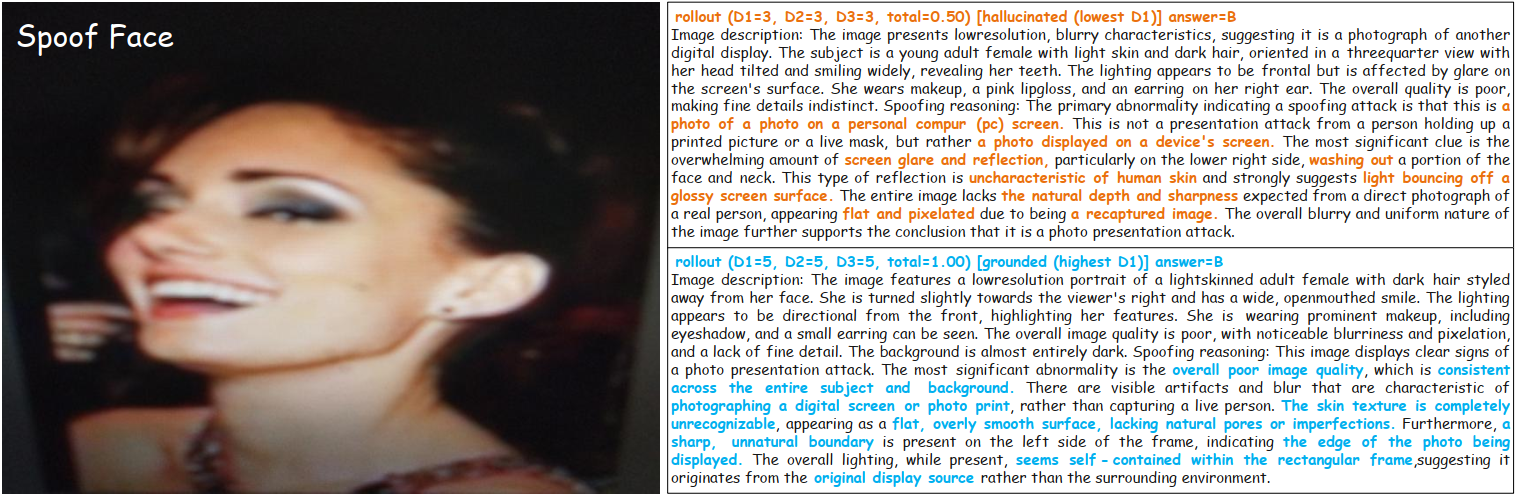}
		\caption{The grounding-aware judge scores the rollouts sampled for one image on three dimensions. The judge assigns a low D1 to a hallucinated chain and a high D1 to a grounded chain, providing a dense process signal that the binary verdict alone cannot.}
		\label{fig:rollout}
	\end{figure}
	
	Figure~\ref{fig:rollout} gives a concrete view of why the judge reward is the active ingredient. For a single image, the policy samples eight rollouts. Several of them arrive at the same verdict through visibly different reasoning: some cite artifacts that are not present in the image, while others describe evidence that is actually there. The binary verdict cannot distinguish these cases, since all of them receive the same accuracy reward, whereas the judge assigns a markedly lower D1 to the hallucinated chains. The group-normalized advantage in Eq.~\eqref{eq:adv} then turns this dense, process-level difference into a learning signal, which is exactly the supervision that a verdict-only reward cannot supply.
	
	\section{Discussion}
	\label{sec:discussion}
	\subsection{Why the Judge Reward Changes Behavior}
	Why does the grounding-aware judge reward change the model's behavior? FAS reasoning has no rule-verifiable ground truth; the only verifiable signal from the label is the binary verdict, which is sparse and invites reward hacking. The grounding-aware judge supplies a denser signal: it rewards reasoning chains anchored in the image evidence rather than in memorized shortcuts. Physical artifacts are verbalizable and image-grounded, so the judge can verify grounding and steer the model toward generalizable forensic reasoning. The qualitative evidence supports this account: on 94 live samples where SFT falsely rejects and RL corrects, SFT hallucinates recapture, print, or gloss artifacts, whereas RL grounds its judgment in the actual image content.
	
	A natural comparison is with recent RL-FAS methods~\cite{ma2026pafas,wang2026fasr1,zhang2026tarfas}, which use rule-based class-consistency or reasoning-consistency rewards. Those rewards check whether the reasoning is internally consistent or matches a template, but they do not check whether the reasoning is grounded in the actual image. A direct quantitative comparison is not feasible because their code and custom training corpora are not publicly released. Our contribution is orthogonal and fully reproducible on public benchmarks: the grounding-aware judge is the first to verify FAS reasoning against image evidence.
	
	\subsection{Hyper-Parameter Sensitivity and Seed Stability}
	\label{sec:hyperseed}
	A concern with any RL stage is that reported gains may be an artifact of a lucky hyper-parameter or a favorable seed. We address both. Table~\ref{tab:hyper} sweeps the learning rate and the KL coefficient around the chosen configuration, and Table~\ref{tab:seed} repeats the best configuration across three random seeds.
	
	\begin{table}[htbp]
		\centering
		\caption{Hyper-parameter sweep (judge weight 0.4 unless noted). Lower ACER is better.}
		\label{tab:hyper}
		\small
		\begin{tabular}{l c c c}
			\toprule
			\textbf{Variant} & \textbf{CelebA-Spoof} & \textbf{LCC-FASD} & \textbf{MSU-MFSD} \\
			& (ID) & (OOD) & (OOD) \\
			\midrule
			lr $2\times10^{-6}$ & 0.034 & 0.276 & 0.423 \\
			lr $5\times10^{-7}$, 2000 steps & 0.033 & 0.272 & 0.411 \\
			$\beta{=}0.04$ & 0.035 & 0.280 & 0.432 \\
			$\beta{=}0.2$ & 0.030 & 0.285 & 0.428 \\
			\midrule
			best (lr $1\times10^{-6}$, $\beta{=}0.1$) & 0.035 & \textbf{0.273} & \textbf{0.409} \\
			\bottomrule
		\end{tabular}
	\end{table}
	
	\textbf{Learning rate.} Raising the learning rate to $2\times10^{-6}$ degrades MSU-MFSD to $0.423$, and halving it while doubling the step budget to 2000 yields $0.411$, close to but slightly worse than the chosen $0.409$. The behavior is asymmetric: an overly large step size hurts the harder domain more than a smaller one does, which is consistent with the judge signal being comparatively fine-grained; large updates overshoot the region in which the policy actually respects the grounding assessment, whereas smaller updates simply converge more slowly without harming the outcome. The chosen $1\times10^{-6}$ sits on the flat side of this trade-off.
	
	\textbf{KL anchor.} The KL coefficient controls how far the policy may drift from the SFT reference. Loosening it to $\beta{=}0.04$ loses the OOD gain on MSU-MFSD ($0.432$), indicating that without a sufficient anchor the policy drifts toward reward-hacking solutions that score well with the judge but are not grounded in the image; the false-rejection bias discussed in the introduction is one such solution. Tightening it to $\beta{=}0.2$, on the other hand, stops the OOD improvement ($0.428$ on MSU-MFSD) while improving in-domain accuracy to $0.030$: the policy is held so close to the SFT reference that it cannot move toward the grounded reasoning the judge rewards. The useful range is therefore narrow and bracketed on both sides, which is itself evidence that the OOD gain comes from a specific behavioral change rather than from generic regularization.
	
	\textbf{Seed stability.} Table~\ref{tab:seed} reports three seeds. All three improve both OOD benchmarks relative to the CoT SFT baseline ($0.288$ on LCC-FASD, $0.431$ on MSU-MFSD), and in-domain accuracy stays stable ($0.034\pm0.003$). The OOD variance is small on LCC-FASD ($0.277\pm0.005$) and larger on MSU-MFSD ($0.413\pm0.012$): the harder domain is not only harder on average but also more variable across runs, which is expected when the supervision signal is a noisy judge rather than an exact rule. Notably, no seed recovers the full $0.409$ of the best run, but every seed stays clearly below the baseline, so the direction of the effect is robust even though its exact size is not. Together with Table~\ref{tab:hyper}, these results confirm that the judge reward, rather than any specific RL hyper-parameter or seed, is the active ingredient.
	
	\begin{table}[htbp]
		\centering
		\caption{ACER across three random seeds (mean $\pm$ std).}
		\label{tab:seed}
		\small
		\begin{tabular}{l c c c}
			\toprule
			\textbf{Seed} & \textbf{CelebA-Spoof} & \textbf{LCC-FASD} & \textbf{MSU-MFSD} \\
			& (ID) & (OOD) & (OOD) \\
			\midrule
			42 & 0.035 & 0.273 & 0.409 \\
			123 & 0.036 & 0.277 & 0.404 \\
			2024 & 0.031 & 0.282 & 0.427 \\
			\midrule
			mean $\pm$ std & $0.034 \pm 0.003$ & $0.277 \pm 0.005$ & $0.413 \pm 0.012$ \\
			\bottomrule
		\end{tabular}
	\end{table}
	
	\subsection{Limitations}
	A limitation is that the observed OOD gain is consistent but limited (LCC-FASD $-0.015$, MSU-MFSD $-0.021$). Because OOD is used as a validation setting rather than the objective, this limited magnitude does not undermine the method's contribution; it delimits what the current judge can achieve. The target domains exhibit substantial shift (the MMD distance from CelebA-Spoof to MSU-MFSD is $0.161$, versus $0.113$ to LCC-FASD), and even CLIP degrades to $0.410$ on MSU-MFSD, so the difficulty is objective rather than an artifact of our training setup. The seed results in Table~\ref{tab:seed} point to the same conclusion from a different angle: the run-to-run variance on MSU-MFSD ($\pm0.012$) is of the same order as the gap to the best run, so the ceiling is set by how reliably the judge can assess grounding on a strongly shifted domain. The limited gain therefore stems from the judge's grounding-capability bottleneck rather than from a wrong pipeline direction, and a stronger judge is expected to bring larger gains.
	
	\section{Conclusion}
	\label{sec:conclusion}
	We proposed a grounding-aware judge reward for RL post-training of a FAS MLLM. The method addresses a structural obstacle in applying RL to FAS: the reasoning has no rule-verifiable ground truth, so a verdict-only reward is sparse and invites shortcut behavior. Our answer is to turn the image itself into the verification signal, with a 32B vision-language judge that scores each reasoning chain on evidence anchoring, logical coherence, and domain knowledge. CoT SFT followed by GRPO with this reward improves out-of-domain generalization on two benchmarks while preserving in-domain accuracy. A full ablation further isolates the judge reward as the ingredient that extends the gain to the harder benchmark. To our knowledge this is the first use of LLM-as-judge process supervision for FAS.



\begin{thebibliography}{26}
\ifx \bisbn   \undefined \def \bisbn  #1{ISBN #1}\fi
\ifx \binits  \undefined \def \binits#1{#1}\fi
\ifx \bauthor  \undefined \def \bauthor#1{#1}\fi
\ifx \batitle  \undefined \def \batitle#1{#1}\fi
\ifx \bjtitle  \undefined \def \bjtitle#1{#1}\fi
\ifx \bvolume  \undefined \def \bvolume#1{\textbf{#1}}\fi
\ifx \byear  \undefined \def \byear#1{#1}\fi
\ifx \bissue  \undefined \def \bissue#1{#1}\fi
\ifx \bfpage  \undefined \def \bfpage#1{#1}\fi
\ifx \blpage  \undefined \def \blpage #1{#1}\fi
\ifx \burl  \undefined \def \burl#1{\textsf{#1}}\fi
\ifx \doiurl  \undefined \def \doiurl#1{\url{https://doi.org/#1}}\fi
\ifx \betal  \undefined \def \betal{\textit{et al.}}\fi
\ifx \binstitute  \undefined \def \binstitute#1{#1}\fi
\ifx \binstitutionaled  \undefined \def \binstitutionaled#1{#1}\fi
\ifx \bctitle  \undefined \def \bctitle#1{#1}\fi
\ifx \beditor  \undefined \def \beditor#1{#1}\fi
\ifx \bpublisher  \undefined \def \bpublisher#1{#1}\fi
\ifx \bbtitle  \undefined \def \bbtitle#1{#1}\fi
\ifx \bedition  \undefined \def \bedition#1{#1}\fi
\ifx \bseriesno  \undefined \def \bseriesno#1{#1}\fi
\ifx \blocation  \undefined \def \blocation#1{#1}\fi
\ifx \bsertitle  \undefined \def \bsertitle#1{#1}\fi
\ifx \bsnm \undefined \def \bsnm#1{#1}\fi
\ifx \bsuffix \undefined \def \bsuffix#1{#1}\fi
\ifx \bparticle \undefined \def \bparticle#1{#1}\fi
\ifx \barticle \undefined \def \barticle#1{#1}\fi
\bibcommenthead
\ifx \bconfdate \undefined \def \bconfdate #1{#1}\fi
\ifx \botherref \undefined \def \botherref #1{#1}\fi
\ifx \url \undefined \def \url#1{\textsf{#1}}\fi
\ifx \bchapter \undefined \def \bchapter#1{#1}\fi
\ifx \bbook \undefined \def \bbook#1{#1}\fi
\ifx \bcomment \undefined \def \bcomment#1{#1}\fi
\ifx \oauthor \undefined \def \oauthor#1{#1}\fi
\ifx \citeauthoryear \undefined \def \citeauthoryear#1{#1}\fi
\ifx \endbibitem  \undefined \def \endbibitem {}\fi
\ifx \bconflocation  \undefined \def \bconflocation#1{#1}\fi
\ifx \arxivurl  \undefined \def \arxivurl#1{\textsf{#1}}\fi
\csname PreBibitemsHook\endcsname

\bibitem[\protect\citeauthoryear{Zhang et~al.}{2020}]{zhang2020celeba}
\begin{bchapter}
\bauthor{\bsnm{Zhang}, \binits{Y.}},
\bauthor{\bsnm{Yin}, \binits{Z.}},
\bauthor{\bsnm{Li}, \binits{Y.}},
\bauthor{\bsnm{Yin}, \binits{G.}},
\bauthor{\bsnm{Yan}, \binits{J.}},
\bauthor{\bsnm{Shao}, \binits{J.}},
\bauthor{\bsnm{Liu}, \binits{Z.}}:
\bctitle{{CelebA-Spoof}: Large-scale face anti-spoofing dataset with rich
  annotations}.
In: \bbtitle{European Conference on Computer Vision (ECCV)},
pp. \bfpage{70}--\blpage{85}
(\byear{2020})
\end{bchapter}
\endbibitem

\bibitem[\protect\citeauthoryear{Liu et~al.}{2018}]{liu2018learning}
\begin{bchapter}
\bauthor{\bsnm{Liu}, \binits{Y.}},
\bauthor{\bsnm{Jourabloo}, \binits{A.}},
\bauthor{\bsnm{Liu}, \binits{X.}}:
\bctitle{Learning deep models for face anti-spoofing: Binary or auxiliary
  supervision}.
In: \bbtitle{IEEE/CVF Conference on Computer Vision and Pattern Recognition
  (CVPR)},
pp. \bfpage{389}--\blpage{398}
(\byear{2018})
\end{bchapter}
\endbibitem

\bibitem[\protect\citeauthoryear{Wen et~al.}{2015}]{wen2015msu}
\begin{barticle}
\bauthor{\bsnm{Wen}, \binits{D.}},
\bauthor{\bsnm{Han}, \binits{H.}},
\bauthor{\bsnm{Jain}, \binits{A.K.}}:
\batitle{Face spoof detection with image distortion analysis}.
\bjtitle{IEEE Transactions on Information Forensics and Security}
\bvolume{10}(\bissue{4}),
\bfpage{746}--\blpage{761}
(\byear{2015})
\end{barticle}
\endbibitem

\bibitem[\protect\citeauthoryear{Yu et~al.}{2023}]{yu2023survey}
\begin{barticle}
\bauthor{\bsnm{Yu}, \binits{Z.}},
\bauthor{\bsnm{Li}, \binits{X.}},
\bauthor{\bsnm{Niu}, \binits{X.}},
\bauthor{\bsnm{Shi}, \binits{J.}},
\bauthor{\bsnm{Zhao}, \binits{G.}}:
\batitle{Deep learning for face anti-spoofing: A survey}.
\bjtitle{IEEE Transactions on Pattern Analysis and Machine Intelligence}
\bvolume{45}(\bissue{5}),
\bfpage{5609}--\blpage{5631}
(\byear{2023})
\end{barticle}
\endbibitem

\bibitem[\protect\citeauthoryear{Team}{2025}]{qwen2025vl}
\begin{botherref}
\oauthor{\bsnm{Team}, \binits{Q.}}:
Qwen2.5-VL Technical Report
(2025)
\end{botherref}
\endbibitem

\bibitem[\protect\citeauthoryear{Radford et~al.}{2021}]{radford2021clip}
\begin{bchapter}
\bauthor{\bsnm{Radford}, \binits{A.}},
\bauthor{\bsnm{Kim}, \binits{J.W.}},
\bauthor{\bsnm{Hallacy}, \binits{C.}},
\bauthor{\bsnm{Ramesh}, \binits{A.}},
\bauthor{\bsnm{Goh}, \binits{G.}},
\bauthor{\bsnm{Agarwal}, \binits{S.}},
\bauthor{\bsnm{Sastry}, \binits{G.}},
\bauthor{\bsnm{Askell}, \binits{A.}},
\bauthor{\bsnm{Mishkin}, \binits{P.}},
\bauthor{\bsnm{Clark}, \binits{J.}}, \betal:
\bctitle{Learning transferable visual models from natural language
  supervision}.
In: \bbtitle{International Conference on Machine Learning (ICML)}
(\byear{2021})
\end{bchapter}
\endbibitem

\bibitem[\protect\citeauthoryear{Jiang et~al.}{2025}]{jiang2025ms}
\begin{bchapter}
\bauthor{\bsnm{Jiang}, \binits{N.}},
\bauthor{\bsnm{Zeng}, \binits{D.}}, \betal:
\bctitle{{MS-UFAD}: A large-scale dataset for real-world unified face attack
  detection with text descriptions}.
In: \bbtitle{IEEE International Conference on Acoustics, Speech and Signal
  Processing (ICASSP)},
pp. \bfpage{10711}--\blpage{10715}
(\byear{2025})
\end{bchapter}
\endbibitem

\bibitem[\protect\citeauthoryear{He et~al.}{2024}]{he2024joint}
\begin{bchapter}
\bauthor{\bsnm{He}, \binits{X.}},
\bauthor{\bsnm{Liang}, \binits{D.}},
\bauthor{\bsnm{Yang}, \binits{S.}},
\bauthor{\bsnm{Hao}, \binits{Z.}},
\bauthor{\bsnm{Ma}, \binits{H.}},
\bauthor{\bsnm{Mao}, \binits{B.}},
\bauthor{\bsnm{Li}, \binits{X.}},
\bauthor{\bsnm{Wang}, \binits{Y.}},
\bauthor{\bsnm{Yan}, \binits{P.}},
\bauthor{\bsnm{Liu}, \binits{A.}}:
\bctitle{Joint physical-digital facial attack detection via simulating spoofing
  clues}.
In: \bbtitle{IEEE/CVF Conference on Computer Vision and Pattern Recognition
  Workshops (CVPRW)},
pp. \bfpage{995}--\blpage{1004}
(\byear{2024})
\end{bchapter}
\endbibitem

\bibitem[\protect\citeauthoryear{Ma et~al.}{2026}]{ma2026pafas}
\begin{bchapter}
\bauthor{\bsnm{Ma}, \binits{Y.}},
\bauthor{\bsnm{Lin}, \binits{X.}},
\bauthor{\bsnm{Xu}, \binits{Y.}},
\bauthor{\bsnm{Xie}, \binits{W.}},
\bauthor{\bsnm{Yu}, \binits{Z.}}:
\bctitle{{PA-FAS}: Towards interpretable and generalizable multimodal face
  anti-spoofing via path-augmented reinforcement learning}.
In: \bbtitle{Proceedings of the AAAI Conference on Artificial Intelligence},
vol. \bseriesno{40},
pp. \bfpage{7856}--\blpage{7864}
(\byear{2026}).
\doiurl{10.1609/aaai.v40i10.37729}
\end{bchapter}
\endbibitem

\bibitem[\protect\citeauthoryear{Ouyang et~al.}{2022}]{ouyang2022instructgpt}
\begin{bchapter}
\bauthor{\bsnm{Ouyang}, \binits{L.}},
\bauthor{\bsnm{Wu}, \binits{J.}},
\bauthor{\bsnm{Jiang}, \binits{X.}},
\bauthor{\bsnm{Almeida}, \binits{D.}},
\bauthor{\bsnm{Wainwright}, \binits{C.L.}},
\bauthor{\bsnm{Mishkin}, \binits{P.}},
\bauthor{\bsnm{Zhang}, \binits{C.}},
\bauthor{\bsnm{Agarwal}, \binits{S.}},
\bauthor{\bsnm{Slama}, \binits{K.}},
\bauthor{\bsnm{Ray}, \binits{A.}}, \betal:
\bctitle{Training language models to follow instructions with human feedback}.
In: \bbtitle{Advances in Neural Information Processing Systems (NeurIPS)}
(\byear{2022})
\end{bchapter}
\endbibitem

\bibitem[\protect\citeauthoryear{Lightman et~al.}{2023}]{lightman2023let}
\begin{botherref}
\oauthor{\bsnm{Lightman}, \binits{H.}}, et al.:
Let's Verify Step by Step
(2023)
\end{botherref}
\endbibitem

\bibitem[\protect\citeauthoryear{Shao et~al.}{2024}]{shao2024deepseekmath}
\begin{botherref}
\oauthor{\bsnm{Shao}, \binits{Z.}},
\oauthor{\bsnm{Wang}, \binits{P.}},
\oauthor{\bsnm{Zhu}, \binits{Q.}},
\oauthor{\bsnm{Xu}, \binits{R.}},
\oauthor{\bsnm{Song}, \binits{J.}},
\oauthor{\bsnm{Bi}, \binits{X.}},
\oauthor{\bsnm{Zhang}, \binits{H.}},
\oauthor{\bsnm{Zhang}, \binits{M.}},
\oauthor{\bsnm{Li}, \binits{Y.K.}},
\oauthor{\bsnm{Wu}, \binits{Y.}},
\oauthor{\bsnm{Guo}, \binits{D.}}:
{DeepSeekMath}: Pushing the Limits of Mathematical Reasoning in Open Language
  Models
(2024)
\end{botherref}
\endbibitem

\bibitem[\protect\citeauthoryear{Wang et~al.}{2026}]{wang2026fasr1}
\begin{botherref}
\oauthor{\bsnm{Wang}, \binits{H.}}, et al.:
{FAS-R1}: A Unified Multi-Task {MLLM} for Reasoning Face Anti-Spoofing
(2026)
\end{botherref}
\endbibitem

\bibitem[\protect\citeauthoryear{Zhang et~al.}{2026}]{zhang2026tarfas}
\begin{bchapter}
\bauthor{\bsnm{Zhang}, \binits{H.}},
\bauthor{\bsnm{Wang}, \binits{K.}},
\bauthor{\bsnm{Zhang}, \binits{G.}},
\bauthor{\bsnm{Yue}, \binits{H.}},
\bauthor{\bsnm{Tan}, \binits{Z.}},
\bauthor{\bsnm{Peng}, \binits{S.}},
\bauthor{\bsnm{Zhang}, \binits{T.}},
\bauthor{\bsnm{Tan}, \binits{X.}},
\bauthor{\bsnm{Chen}, \binits{K.}},
\bauthor{\bsnm{He}, \binits{W.}},
\bauthor{\bsnm{Wang}, \binits{J.}},
\bauthor{\bsnm{Liu}, \binits{A.}},
\bauthor{\bsnm{Zhu}, \binits{X.}},
\bauthor{\bsnm{Lei}, \binits{Z.}}:
\bctitle{From intuition to investigation: A tool-augmented reasoning {MLLM}
  framework for generalizable face anti-spoofing}.
In: \bbtitle{IEEE/CVF Conference on Computer Vision and Pattern Recognition
  (CVPR)}
(\byear{2026})
\end{bchapter}
\endbibitem

\bibitem[\protect\citeauthoryear{Sun et~al.}{2021}]{sun2021understanding}
\begin{barticle}
\bauthor{\bsnm{Sun}, \binits{Y.}},
\bauthor{\bsnm{Xiong}, \binits{H.}},
\bauthor{\bsnm{Yiu}, \binits{S.M.}}:
\batitle{Understanding deep face anti-spoofing: From the perspective of data}.
\bjtitle{The Visual Computer}
\bvolume{37}(\bissue{5}),
\bfpage{1015}--\blpage{1028}
(\byear{2021})
\doiurl{10.1007/s00371-020-01849-x}
\end{barticle}
\endbibitem

\bibitem[\protect\citeauthoryear{He et~al.}{2023}]{he2023lightweight}
\begin{barticle}
\bauthor{\bsnm{He}, \binits{D.}},
\bauthor{\bsnm{He}, \binits{X.}},
\bauthor{\bsnm{Yuan}, \binits{R.}},
\bauthor{\bsnm{Li}, \binits{Y.}},
\bauthor{\bsnm{Shen}, \binits{C.}}:
\batitle{Lightweight network-based multi-modal feature fusion for face
  anti-spoofing}.
\bjtitle{The Visual Computer}
\bvolume{39}(\bissue{4}),
\bfpage{1423}--\blpage{1435}
(\byear{2023})
\doiurl{10.1007/s00371-022-02420-6}
\end{barticle}
\endbibitem

\bibitem[\protect\citeauthoryear{Tobin et~al.}{2017}]{tobin2017domain}
\begin{bchapter}
\bauthor{\bsnm{Tobin}, \binits{J.}},
\bauthor{\bsnm{Fong}, \binits{R.}},
\bauthor{\bsnm{Ray}, \binits{A.}},
\bauthor{\bsnm{Schneider}, \binits{J.}},
\bauthor{\bsnm{Zaremba}, \binits{W.}},
\bauthor{\bsnm{Abbeel}, \binits{P.}}:
\bctitle{Domain randomization for transferring deep neural networks from
  simulation to the real world}.
In: \bbtitle{IEEE/RSJ International Conference on Intelligent Robots and
  Systems (IROS)}
(\byear{2017})
\end{bchapter}
\endbibitem

\bibitem[\protect\citeauthoryear{Liu et~al.}{2024}]{liu2024cfpl}
\begin{bchapter}
\bauthor{\bsnm{Liu}, \binits{A.}}, \betal:
\bctitle{Class free prompt learning for generalizable face anti-spoofing}.
In: \bbtitle{IEEE/CVF Conference on Computer Vision and Pattern Recognition
  (CVPR)}
(\byear{2024})
\end{bchapter}
\endbibitem

\bibitem[\protect\citeauthoryear{Li et~al.}{2025}]{li2025optimal}
\begin{bchapter}
\bauthor{\bsnm{Li}, \binits{Z.}},
\bauthor{\bsnm{Zhao}, \binits{T.}},
\bauthor{\bsnm{Xu}, \binits{X.}},
\bauthor{\bsnm{Zhang}, \binits{Z.}},
\bauthor{\bsnm{Li}, \binits{Z.}},
\bauthor{\bsnm{Chen}, \binits{X.}},
\bauthor{\bsnm{Zhang}, \binits{Q.}},
\bauthor{\bsnm{Bergamo}, \binits{A.}},
\bauthor{\bsnm{Jain}, \binits{A.K.}},
\bauthor{\bsnm{Xing}, \binits{Y.}}:
\bctitle{Optimal transport-guided source-free adaptation for face
  anti-spoofing}.
In: \bbtitle{IEEE/CVF Conference on Computer Vision and Pattern Recognition
  (CVPR)},
pp. \bfpage{24351}--\blpage{24363}
(\byear{2025})
\end{bchapter}
\endbibitem

\bibitem[\protect\citeauthoryear{Wang et~al.}{2026}]{wang2026faceshield}
\begin{bchapter}
\bauthor{\bsnm{Wang}, \binits{H.}},
\bauthor{\bsnm{Shi}, \binits{Y.}},
\bauthor{\bsnm{Tao}, \binits{Z.}},
\bauthor{\bsnm{Gao}, \binits{Y.}},
\bauthor{\bsnm{Zhang}, \binits{L.}},
\bauthor{\bsnm{Lin}, \binits{X.}},
\bauthor{\bsnm{Feng}, \binits{J.}},
\bauthor{\bsnm{Yuan}, \binits{X.}},
\bauthor{\bsnm{Yu}, \binits{Z.}},
\bauthor{\bsnm{Cao}, \binits{X.}}:
\bctitle{{FaceShield}: Explainable face anti-spoofing with multimodal large
  language models}.
In: \bbtitle{Proceedings of the AAAI Conference on Artificial Intelligence},
vol. \bseriesno{40},
pp. \bfpage{9811}--\blpage{9819}
(\byear{2026})
\end{bchapter}
\endbibitem

\bibitem[\protect\citeauthoryear{Zheng et~al.}{2025}]{zheng2025gspo}
\begin{botherref}
\oauthor{\bsnm{Zheng}, \binits{C.}},
\oauthor{\bsnm{Liu}, \binits{S.}},
\oauthor{\bsnm{Li}, \binits{M.}},
\oauthor{\bsnm{Chen}, \binits{X.-H.}},
\oauthor{\bsnm{Yu}, \binits{B.}},
\oauthor{\bsnm{Gao}, \binits{C.}},
\oauthor{\bsnm{Dang}, \binits{K.}},
\oauthor{\bsnm{Liu}, \binits{Y.}},
\oauthor{\bsnm{Men}, \binits{R.}},
\oauthor{\bsnm{Yang}, \binits{A.}},
\oauthor{\bsnm{Zhou}, \binits{J.}},
\oauthor{\bsnm{Lin}, \binits{J.}}:
Group Sequence Policy Optimization
(2025)
\end{botherref}
\endbibitem

\bibitem[\protect\citeauthoryear{Zheng et~al.}{2023}]{zheng2023judge}
\begin{bchapter}
\bauthor{\bsnm{Zheng}, \binits{L.}},
\bauthor{\bsnm{Chiang}, \binits{W.-L.}},
\bauthor{\bsnm{Sheng}, \binits{Y.}}, \betal:
\bctitle{Judging {LLM}-as-a-{J}udge with {MT-Bench} and {Chatbot Arena}}.
In: \bbtitle{Advances in Neural Information Processing Systems (NeurIPS)}
(\byear{2023})
\end{bchapter}
\endbibitem

\bibitem[\protect\citeauthoryear{Yu et~al.}{2025}]{yu2025judge}
\begin{bchapter}
\bauthor{\bsnm{Yu}, \binits{J.}},
\bauthor{\bsnm{Sun}, \binits{S.}},
\bauthor{\bsnm{Hu}, \binits{X.}},
\bauthor{\bsnm{Yan}, \binits{J.}},
\bauthor{\bsnm{Yu}, \binits{K.}},
\bauthor{\bsnm{Li}, \binits{X.}}:
\bctitle{Improve {LLM}-as-a-{J}udge ability as a general ability}.
In: \bbtitle{Proceedings of EMNLP},
pp. \bfpage{14099}--\blpage{14115}
(\byear{2025})
\end{bchapter}
\endbibitem

\bibitem[\protect\citeauthoryear{Jiang et~al.}{2025}]{jiang2025exploring}
\begin{barticle}
\bauthor{\bsnm{Jiang}, \binits{F.}},
\bauthor{\bsnm{Li}, \binits{Q.}},
\bauthor{\bsnm{Wang}, \binits{W.}},
\bauthor{\bsnm{Wang}, \binits{G.}},
\bauthor{\bsnm{Liu}, \binits{B.}},
\bauthor{\bsnm{Sun}, \binits{Z.}}:
\batitle{Exploring task-solving paradigm for generalized cross-domain face
  anti-spoofing via reinforcement fine-tuning}.
\bjtitle{arXiv preprint arXiv:2506.21895}
(\byear{2025})
\doiurl{10.48550/arXiv.2506.21895}
\end{barticle}
\endbibitem

\bibitem[\protect\citeauthoryear{Yu et~al.}{2026}]{yu2026msmfad}
\begin{barticle}
\bauthor{\bsnm{Yu}, \binits{X.}},
\bauthor{\bsnm{Li}, \binits{R.}},
\bauthor{\bsnm{Shi}, \binits{S.}},
\bauthor{\bsnm{You}, \binits{X.}}:
\batitle{{MS-MFAD}: Multimodal large language models for face anti-spoofing
  detection}.
\bjtitle{arXiv preprint arXiv:2608.17328}
(\byear{2026})
\doiurl{10.48550/arXiv.2608.17328}
\end{barticle}
\endbibitem

\bibitem[\protect\citeauthoryear{Hu et~al.}{2021}]{hu2021lora}
\begin{botherref}
\oauthor{\bsnm{Hu}, \binits{E.J.}},
\oauthor{\bsnm{Shen}, \binits{Y.}},
\oauthor{\bsnm{Wallis}, \binits{P.}},
\oauthor{\bsnm{Allen-Zhu}, \binits{Z.}},
\oauthor{\bsnm{Li}, \binits{Y.}},
\oauthor{\bsnm{Wang}, \binits{S.}},
\oauthor{\bsnm{Wang}, \binits{L.}},
\oauthor{\bsnm{Chen}, \binits{W.}}:
{LoRA}: Low-Rank Adaptation of Large Language Models
(2021)
\end{botherref}
\endbibitem

\end{thebibliography}
\end{document}